\documentclass[a4paper]{article}
\usepackage{authblk} 

\usepackage[english]{babel}
\usepackage[utf8x]{inputenc}
\usepackage[T1]{fontenc}

\usepackage[a4paper,top=3cm,bottom=2cm,left=3cm,right=3cm,marginparwidth=1.75cm]{geometry}

\usepackage{amsmath}
\usepackage{verbatim}
\usepackage{graphicx}
\usepackage{algorithm}
\usepackage{algpseudocode}
\usepackage{diagbox}
\usepackage{enumitem}
\usepackage{wrapfig}
\usepackage{multirow}
\usepackage{indentfirst}
\usepackage{pifont}
\newcommand{\cmark}{\ding{51}}%
\newcommand{\xmark}{\ding{55}}%
\usepackage[colorlinks=true, allcolors=blue]{hyperref}

\title{Overview of TREC 2024 Medical Video Question Answering (MedVidQA) Track
}

\author{Deepak Gupta}
\author{Dina Demner-Fushman}
\affil{National Library of Medicine, NIH}

\date{}
\begin{document}
\maketitle
\section{Overview} \label{sec:into}
One of the key goals of artificial intelligence (AI) is the development of a multimodal system that facilitates communication with the visual world (image and video) using a natural language query. Earlier works \cite{lau2018dataset,yadav2021reinforcement,yadav2022question,voorhees2021trec, yadav2022towards,yadav2023towards}, on medical question answering primarily focused on textual and visual (image) modalities, which may be inefficient in answering questions requiring demonstration. In recent years, significant progress has been achieved due to the introduction of large-scale language-vision datasets and the development of efficient deep neural techniques that bridge the gap between language and visual understanding. Improvements have been made in numerous vision-and-language tasks, such as visual captioning \cite{li2020oscar,luo2020univl}, visual question answering \cite{zhang2021vinvl}, and natural language video localization \cite{anne2017localizing}. Most of the existing work on language vision focused on creating datasets and developing solutions for open-domain applications. We believe medical videos may provide the best possible answers to many first aid, medical emergency, and medical education questions. With increasing interest in AI to support clinical decision-making and improve patient engagement \cite{hhsAI}, there is a need to explore such challenges and develop efficient algorithms for medical language-video understanding and generation. Toward this, we introduced new tasks to foster research toward designing systems that can understand medical videos to provide visual answers to natural language questions, and are equipped with multimodal capability to generate instruction steps from the medical video. These tasks have the potential to support the development of sophisticated downstream applications that can benefit the public and medical professionals.

\section{Tasks} \label{sec:task}
\begin{itemize}[noitemsep]
    \item \textbf{Task A: Video Corpus Visual Answer Localization (VCVAL).} 
    Given a medical query and a collection of videos, the task aims to retrieve the appropriate video from the video collection and then locate the temporal segments (start and end timestamps) in the video where the answer to the medical query is shown or the explanation is illustrated in the video. The proposed VCVAL task can be considered as video retrieval and then find a series of ``\textit{medical instructional activity-based frame localization}'' where a potential solution first searches for all medical instructional activities for a given medical query and then locates the activities in an untrimmed medical instructional video. This task is the extension of the MVAL task introduced in MedVidQA-2022 \cite{gupta2022overview}, where we only focused on locating the segment from a given video. In contrast, the VCVAL task deals with relevant video retrieval followed by the visual answer segment localization (\textit{cf.} Figure \ref{fig:VCCAL-task}). The video retrieval system requires the ability to identify the medical instructional video and retrieve the most relevant video for the health-related query.
\begin{figure}[h]
    \centering
    \includegraphics[width=\linewidth]{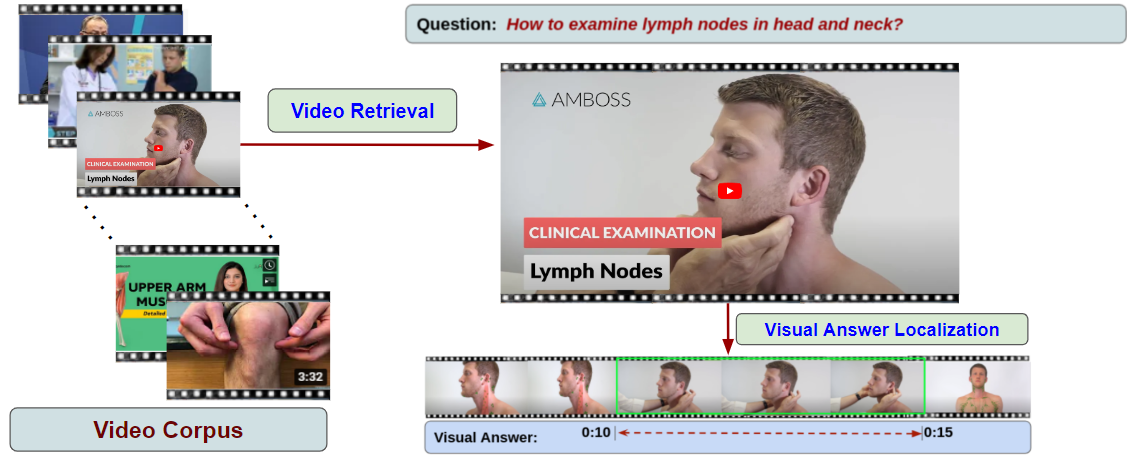}
    \caption{Visualization of the proposed video corpus visual answer localization (VCVAL) task. The VCVAL task consists of two sub-tasks: video retrieval and visual answer localization.}
    \label{fig:VCCAL-task}
\end{figure}

\begin{figure}[h]
    \centering
    \includegraphics[width=\linewidth]{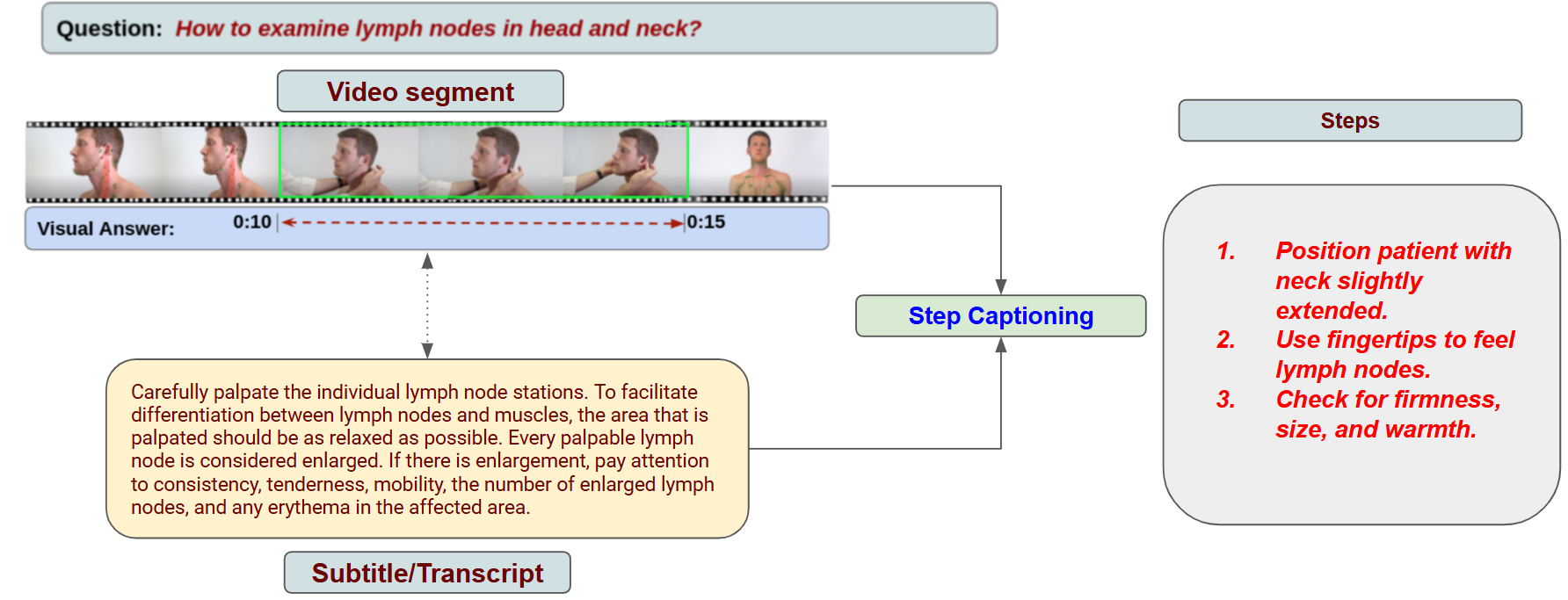}
    \caption{Visualization of the proposed query-focused instructional step captioning (QFISC) task.}
    \label{fig:QFISC-task}
\end{figure}

      \item \textbf{Task B: Query-Focused Instructional Step Captioning (QFISC). } 
Given a medical query and a video, this task aims to generate
step-by-step textual summaries of the visual instructional segment that can be considered as the answer to the medical query. The proposed QFISC task can be considered an extension of the visual answer localization task, where the system needs to locate a series of instructional segments that serve as the answer to the query. The QFISC requires identifying the boundaries of instructional steps and generating a caption for each step. This task comes under multimodal generation, where the system has to consider the video (visual) and subtitle (language) modality to generate (\textit{cf.} Figure \ref{fig:QFISC-task}) the natural language caption. The applications of the QFISC task include the accessibility of video with the textual query, easy indexing of the videos with corresponding step caption, multimodal retrieval, etc.
\end{itemize}

\section{Datasets}
\subsection{VCVAL}
The VCVAL task comprises two subtasks: video retrieval and visual answer localization. For the video retrieval, we developed a video corpus considering the videos from `\textit{Personal Care and Style},' `\textit{Health},' and `\textit{Sports and Fitness}' categories within the HowTo100M \cite{miech2019howto100m} dataset. We follow the strategy discussed in \cite{gupta2024towards} to select the medical instructional videos from the HowTo100M dataset. We also added the video corpus of size $12,657$ released for the TRECVID 2023 MedVidQA track \cite{awad2023trecvid}. This process yielded a total of $48,605$ medical instructional videos, which we considered as a video corpus to retrieve the relevant videos against the query.
To facilitate training and validation of the visual answer localization system, we provided MedVidQA collections \cite{gupta2022dataset} consisting of $3,010$ human-annotated instructional questions and visual answers from $899$ health-related videos. 

We sampled a total of $150$ distinct videos on different topics from the video corpus and created $52$ medical instructional questions following the guidelines discussed in \cite{gupta2022dataset}. In particular, we follow the following guidelines:
\begin{itemize}[noitemsep]
    \item The question is looking for an instructional answer, and
\item The answer should be shown or the explanation must illustrated in the video for the created question, and
\item A video snippet is necessary to answer the question, and
\item The answer should not be given as text (e.g. definitional question) or spoken information without visual aid.
\end{itemize}
We consider the created questions as our test set for the VCVAL task. The created test questions are shown in Table \ref{tab:vcval-test-questions}.

\begin{table}[]
\resizebox{\columnwidth}{!}{%
\begin{tabular}{l|l}
\textbf{QId} & \textbf{Question} \\ \hline
Q1 & How to assess for a dislocated hip through physical examination? \\
Q2 & How to perform a simple exercise to relieve tension in the neck muscles? \\
Q3 & How to get did of toenail fungus? \\
Q4 & How to properly clean ear canal? \\
Q5 & How to properly use a nebulizer? \\
Q6 & How to reduce bleeding and inflammation in the gums? \\
Q7 & How to apply heat or cold therapy to alleviate sciatica pain in the lower back? \\
Q8 & How to handle a groin strain using exercise? \\
Q9 & How to empty a urinary catheter bag safely and properly? \\
Q10 & How to help a person in removing their prosthetic limb? \\
Q11 & How to perform rescue breathing on an infant with a tracheal tube? \\
Q12 & How to use an inhaler with a spacer? \\
Q13 & How to prepare hydrogen peroxide solution for contact lenses? \\
Q14 & How should contact lenses be correctly inserted? \\
Q15 & How to use diaphragmatic breathing to manage asthma? \\
Q16 & How can ice be used to treat calf tendonitis? \\
Q17 & How do you use a water flosser to remove tonsil stones? \\
Q18 & How to floss between an implant and the gum line? \\
Q19 & How do you use a heat pack to help a stiff jaw? \\
Q20 & How do you clean around wisdom teeth? \\
Q21 & How to stop gums from bleeding? \\
Q22 & How do you clear your nose with a nasal spray? \\
Q23 & How to do leg squats to relieve a herniated disc? \\
Q24 & How to stand up from a seated position after ACL reconstruction surgery? \\
Q25 & How to do a self-exam on the lymph nodes in the neck? \\
Q26 & How to do upright row-to-plank exercises for rotator cuff? \\
Q27 & How to relieve pain from a pinched nerve in the neck? \\
Q28 & How to brush your teeth with an electric toothbrush? \\
Q29 & How to brush your teeth if you have braces? \\
Q30 & How can I remove stains from my teeth? \\
Q31 & How to clean your teeth with a handheld pressure washer? \\
Q32 & How is the knee checked for arthritis? \\
Q33 & How do you get rid of scar tissue in the foot? \\
Q34 & How to stretch hip muscles to relieve hip pain? \\
Q35 & How to massage around the ear to lessen jaw pain? \\
Q36 & How do you stretch hamstrings for pain relief? \\
Q37 & How to stretch to ease lower back pain from a pinched nerve? \\
Q38 & How are squats done for relief from knee pain? \\
Q39 & How can cracked feet be treated? \\
Q40 & How is heat therapy used to treat stiff joints? \\
Q41 & How is tape applied to the bicep for pain relief? \\
Q42 & How do you self-massage the jaw for TMJ pain relief? \\
Q43 & How is neck flexion measured? \\
Q44 & How is CPR performed on a child? \\
Q45 & How do you test for a torn Achilles tendon? \\
Q46 & How do you check for a tight jaw? \\
Q47 & How to do hamstring exercises to reduce knee swelling? \\
Q48 & How to use a defibrillator for chest compressions? \\
Q49 & How are metacarpal flexion and extension measured? \\
Q50 & How do you ascend stairs with crutches? \\
Q51 & How to do a glucagon injection? \\
Q52 & How do you stop bleeding with a first aid kit? \\ \hline
\end{tabular}%
}
\caption{Test question for the Video Corpus Visual Answer Localization task.}
\label{tab:vcval-test-questions}
\end{table}
\subsection{QFISC} 
 We utilized the open domain HIREST dataset \cite{zala2023hierarchical} to train the system for the QFISC task. HIREST comprises 3.4K text-video pairs sourced from an instructional video dataset. Among these, 1.1K videos are annotated with moment spans pertinent to text queries. Each moment is further dissected into key instructional steps, complete with captions and timestamps, resulting in a total of 8.6K step captions.

To create a test collection, we used the manually annotated visual segments of the VCVAL task from TRECVID 2023 MedVidQA track \cite{awad2023trecvid}. 
We sampled $140$ visual segments considering the distinct topics from the ground-truth collection of the TRECVID 2023 MedVidQA-VCVAL task and formulated a $90$ visual segment-steps pair dataset. We provided the guidelines shown in Fig. \ref{fig:qfisc-step-caption-guidelines} to the annotators to formulate the step captions. By following the guidelines, we created $90$ visual segment-steps pair which was provided to the participants as test segments to generate the step captions. 
\begin{figure}[!t]
\fbox{\begin{minipage}{0.9\textwidth}
\textbf{Objective}: \textbf{1)} to create step-by-step, well-formed steps of the visual instructional segment that can be considered the answer to the medical query, and \textbf{2)} provide the time stamps (start and end) where the step is being shown or the explanation is illustrated in the video.

Please note the following while creating the steps:
\begin{itemize}[noitemsep]
    \item A video can have multiple visual answer segments.
    \item For each visual answer segment, you need to create instructional steps. 
    \item You can start writing a step with the action verb.

\end{itemize}

\textbf{Well-formed Steps:}
We define a step to be a well-formed natural language step if it satisfies the following:
\begin{enumerate}[noitemsep]
    \item The step does not contain spelling errors. 
    \item A step should not be longer than seven (7) words.
    \item 
Each step should clearly signify an action being demonstrated in the visual segment. 
\item It should be a simple step focusing on a single action. Please avoid writing two actions in a single step. 
\end{enumerate}

\textbf{Examples:}
\begin{itemize}[noitemsep]
    \item Stabilize the arm with the board \cmark
    \item 
Tie the elbow to the board \cmark
\item 
Stabilize the arm and tie the elbow with the board \xmark
\end{itemize}
\textbf{Time-stamps:}
A  timestamp is a way to link to a specific step in the video. A time stamp needs to be entered in MM: SS format. E.g., 02:30 (2 min 30 sec), 10:56 (10 min 56 sec)

\end{minipage}}
\caption{Annotation Guidelines for manual step captioning of QFISC task.}
 \label{fig:qfisc-step-caption-guidelines}
\end{figure}

\section{Judgments} 
\subsection{VCVAL}
The participants were asked to retrieve the relevant videos (up to $1,000$) for each question from the video corpus of having $48,605$ videos. Additionally, the participants also had to provide the start and end timestamps from each retrieved video against a given question, which can be considered a visual answer to the question. In order to judge the relevant videos and corresponding visual answers in the videos, we performed the manual judgments by applying the pooling strategy with pool size=25 (first 10 with probability 1, next 5 with probability 0.3, next 5 with probability 0.2, and next 5 with probability 0.1) of all the submitted videos ($12,184$) and visual answers by the participants. We instructed a total of three assessors with the following guidelines to assess the video:
\paragraph{Objective:} 
\begin{enumerate}[noitemsep]
\item  To judge relevant videos with respect to medical/healthcare instructional questions. A video can be called relevant if it has a visual answer to the question.  
\item For each relevant video, provide the time stamps (start and end) where the answer is being shown or the explanation is illustrated in the video.
\end{enumerate}

\paragraph{Evaluating videos for relevance}:
The videos are judged as being \textit{``Definitely Relevant''}, \textit{``Possibly Relevant''}, or \textit{``Not Relevant''} to the given question. The assessors were presented with videos from the submitted runs. They were instructed to determine if the video was definitely relevant, possibly relevant, or not relevant to the question. In general, a video is definitely relevant if it contains a visual segment that can be considered a complete visual answer to the question. A video can be considered possibly relevant if it contains a visual segment that can be considered a partial/incomplete visual answer to the question. If the visual segments from the videos do not provide any visual answers to the question, the video can be marked as not relevant. The assessors were asked to provide the judgment with the following instructions:

\begin{itemize}[noitemsep]
    \item Only provide the time stamps for definitely relevant and possibly relevant videos.
    \item For each definitely relevant and possibly relevant video, provide the time stamps from the video that can be considered a visual answer.
    \item The time stamps should be the shortest span in the video, which can be considered as a complete (for definitely relevant video) or partially complete (for possibly relevant video) visual answer to the question.
    \item In case a video has multiple visual answers to the same question, assessors were asked to provide all the visual answers.
\end{itemize}

\subsection{QFISC} \label{sec:qfisc-judgement}
We also performed a manual assessment of the participants' submitted step captions. To assess the generated steps, we performed manual judgments of all the steps to the participants' submitted runs. We instructed an assessor with the guidelines provided in Figure \ref{fig:qfisc-step-caption-assessment-guidelines} to assess the system-generated steps.
\begin{figure}[!t]
\fbox{\begin{minipage}{0.9\textwidth}
\textbf{Objective}: To assess the system-generated steps on human evaluation criteria such as completeness, accuracy, and coherency.
The assessor must provide the score on the Likert scale (1-5, 1 signifies least and 5 denotes most) for the system-generated steps using the human evaluation criteria listed below:

\begin{itemize}[noitemsep]
    \item \textbf{Completeness}: Whether the generated steps contain all the necessary steps to perform the certain task to achieve the desired output.
    \item \textbf{Accuracy}: Whether the generated steps are accurate to what is being demonstrated in the visual segment. 
    \item \textbf{Coherency}:  Whether the steps logically connect to each other. The steps should follow the correct ordering for the performed task to achieve the desired output.
\end{itemize}
\end{minipage}}
\caption{Human assessment guidelines for evaluating the system generated steps in QFISC task.}
 \label{fig:qfisc-step-caption-assessment-guidelines}
\end{figure}

\section{Evalaution}
\subsection{Metrics for VCVAL Task}
The VCVAL task consists of two sub-tasks: video retrieval (VR) and visual answer localization (VAL). We evaluated the performance of the video retrieval system in terms of Mean Average Precision (MAP), Recall@k, Precision@k, and nDCG metrics with $k=\{5, 10\}$. We follow the \texttt{trec\_eval}\footnote{\url{https://github.com/usnistgov/trec\_eval}} evaluation library to report the performance of participating systems.

For the VAL task, if the predicted (retrieved by the system) video is from the list of relevant videos (marked by the assessor; we called it ground-truth video), then we compute the overlap between the retrieved and relevant video by the following metrics:
 
\begin{enumerate}
\item  \textbf{Mean Intersection over Union (mIoU): } For a given question $q_i$, IoU is computed as the ratio of intersection area over union area between predicted and ground-truth temporal visual answer segments. It ranges from $0$ to $1$. A larger IoU means the predicted and ground-truth temporal visual answer segments match better, and IoU = 1.0 denotes an exact match. 
The mIoU is defined as the average temporal IoUs for all questions ($N$) in the test set. Formally,
\begin{equation}
    \text{mIoU}=\frac{1}{N}\sum_{i=1}^{i=N} \texttt{IoU}(q_i)
\end{equation}
\item   \textbf{IoU = $\mu$} is another metric used to evaluate the performance of the VAL system. It denotes the percentage of questions for which, out of the top-$n$ retrieved temporal segments, at least one predicted temporal segment having IoU with ground truth is larger than $\mu$.
Formally,
\begin{gather}
    <{R\alpha n,IoU=\mu}> \: =\frac{1}{N}\sum_{i=1}^{i=N} s(q_i, \mu), \: \text{and} \\
    s(q_i, \mu) = 
\begin{cases}
    1 ,& \text{if } IoU(q_i) \geq  \mu\\
    0,              & \text{otherwise}
\end{cases}
\end{gather}
We evaluated the participants' submission by considering $\mu=\{0.3, 0.5, 0.7\}$ and for brevity, we denote the $ <{R\alpha n,IoU=\mu}>$ metric with IoU=$\mu$ $n=\{1,3,5,10\}$
\end{enumerate}

\begin{algorithm}[]
\caption{Align Predicted Steps with Ground Truth Steps}
\label{algo:align_steps}
\textbf{Input:} Predicted steps $P$, Ground truth steps $G$, Overlap threshold $\theta$ \\
\textbf{Output:} True Positives (TP), False Positives (FP), False Negatives (FN)

\begin{algorithmic}[1]
    \State Initialize $TP \gets 0$, $FP \gets 0$, $FN \gets 0$
    \State Set $g\_index \gets 0$ \Comment{Start from the first ground truth step}

    \For{\textbf{each} $p\_step$ \textbf{in} $P$} 
        \State $best\_score \gets -1$, $best\_g\_index \gets -1$ \Comment{Track highest alignment score and index}

        \For{\textbf{each} $g\_step$ \textbf{in} $G[g\_index: ]$}
            \State $score \gets \text{calculate\_alignment\_score}(p\_step, g\_step)$
            \If{$score > best\_score$ \textbf{and} $score \geq \theta$} 
                \State $best\_score \gets score$
                \State $best\_g\_index \gets g\_index$
            \EndIf
        \EndFor

        \If{$best\_g\_index \neq -1$}
            \State $TP \gets TP + 1$
        
            \State $g\_index \gets best\_g\_index + 1$ \Comment{Advance to next ground truth step}
        \Else
            \State $FP \gets FP + 1$ \Comment{No match found for this predicted step}
        \EndIf
    \EndFor

    \State $FN \gets |G| - TP$ \Comment{Count remaining ground truth steps as False Negatives}

    \State \Return $TP$, $FP$, $FN$
\end{algorithmic}
\end{algorithm}

\begin{algorithm}[H]
\caption{Calculate Alignment Score}
\label{algo:alignment_score}
\textbf{Input:} Predicted step $pred\_step$, Ground truth step $gt\_step$, Overlap weight $\alpha=0.5$, ROUGE weight $\beta=0.5$ \\
\textbf{Output:} Alignment score $S$

\begin{algorithmic}[1]
    \State $overlap \gets \text{calculate\_time\_overlap}(pred\_step, gt\_step)$
    \Comment{Compute time overlap between predicted and ground truth steps}
    
    \State $rouge\_score \gets \text{rouge.compute}($
    \State \hspace{2em} $predictions = [pred\_step[\text{``caption''}]],$
    \State \hspace{2em} $references = [gt\_step[\text{``caption''}]])$
    \Comment{Calculate ROUGE-L score for captions}
    
    \State $S \gets \alpha \times overlap + \beta \times rouge\_score[\text{``rougeL''}]$
    \Comment{Weighted combination of overlap and ROUGE-L}

    \State \Return $S$
\end{algorithmic}
\end{algorithm}

\subsection{Metrics for QFISC Task} \label{sec:qfisc-evaluation}
We evaluated the performance of the step captioning task on two fronts: \textbf{(1)} how close the system-generated step caption is to the ground truth step captions, and \textbf{(2)} how well the predicted step segment aligns with the ground truth step segment. We measure the closeness in two ways:  
\begin{enumerate}
    \item With the help of predicted timestamps and sentence level similarity (ROUGE-L) of the step, we align (\textit{c.f.} Algo \ref{algo:align_steps},  \ref{algo:alignment_score}) a generated step to one of the ground truth steps. Towards this, we define the following:
    \begin{itemize}
        \item  TP (True Positives): Represents the count of predicted steps that are present in the ground truth steps.
 \item  FP (False Positives) :
 Represents the count of predicted steps that are not present in the ground truth steps.

 \item  FN (False Negatives): 
  Represents the count of ground truth steps that are not present in the predicted steps.

    \end{itemize}
 Following this, we compute the precision, recall, and f-score. 
    \item We use the \textit{n-gram} matching metrics: BLEU \cite{papineni2002bleu}, ROUGE \cite{lin2004rouge}, METEOR \cite{banerjee2005meteor} and SPICE \cite{anderson2016spice}. Additionally, we use sentence-level embedding-based metrics: BERTScore \cite{zhang2019bertscore} as they capture the semantic similarity between the generated and ground truth captions. All the n-gram metrics are computed between predicted and ground-truth step captions.
\end{enumerate}
To compute the alignment between the predicted step segments and the ground truth step segments, we use the intersection over union (IoU) metric. For a given step, IoU is computed as the ratio of the common segment to the union between the predicted and ground-truth segments. It ranges from $0$ to $1$. For the shorter step, where the step segment lasts, say, only 1-2 seconds, if the system-generated step segment does not match with the ground-truth segment, the system may end up with IoU=0. To deal with such a situation, we will use relaxed IoU, where we extend the segments by $\lambda$ before computing the IoU. We compute the mean of the IoU for all the segments in the test set.

\section{Participating Teams}
We use the NIST-provided Evalbase platform\footnote{\url{https://ir.nist.gov/evalbase}} to release the
datasets, registration, and submissions of the participating teams. In total, $7$ teams participated in the MedVidQA track and submitted $17$ individual runs for the tasks. We have provided (\textit{cf.} Table \ref{tab:team_affiliations}) the team name, affiliations, and their participation in VCVAL and QFISC tasks.

\begin{table}[]
\centering
\resizebox{0.8\linewidth}{!}{%
\begin{tabular}{l|l|c|c}
\hline
\textbf{Team Name}                 & \textbf{Team Affiliations} & \textbf{VCVAL} &\textbf{QFISC} \\ \hline
TJUMI    &     Tianjin University             & \ding{51}    & \ding{55}       \\ 
PolySmart         &       City University of Hong Kong       & \ding{51}       & \ding{51}      \\ 
NCSU           &    North Carolina State University
              & \ding{51}      & \ding{55}     \\ 
NCstate           &    North Carolina State University
              & \ding{51}      & \ding{55}     \\ 
UNCW           &    University of North Carolina Wilmington
              & \ding{51}      & \ding{55}     \\ 

UNCC           &        University of North Carolina at Charlotte        & \ding{51}     & \ding{55}       \\ 
DoshishaUzlDfki &      Doshisha University and University of Lubeck
         & \ding{55}      & \ding{51}       \\ \hline
\end{tabular}%
}
\caption{MedVidQA: Participating teams and their task participation at TREC 2024.} 
\label{tab:team_affiliations}
\end{table}

\begin{table}[]
\resizebox{\columnwidth}{!}{%
\begin{tabular}{l|l|cccccc}
\hline
\textbf{Team} & \textbf{RunID} & \textbf{MAP} & \textbf{R@5} & \textbf{R@10} & \textbf{P@5} & \textbf{P@10} & \textbf{nDCG} \\ \hline
\multirow{2}{*}{TJUMI} & run-2 & 0.2164 & 0.1743 & 0.2219 & 0.3731 & 0.2923 & 0.3033 \\ 
 & run-1 & 0.2161 & 0.1767 & 0.2233 & 0.3808 & 0.2962 & 0.3017 \\ \hline
NCstate & Seahawk\_run-1 & 0.4119 & 0.2652 & 0.429 & 0.4808 & 0.4654 & 0.5738 \\ \hline
\multirow{5}{*}{PolySmart} & 3 & 0.1276 & 0.105 & 0.1668 & 0.2846 & 0.2481 & 0.239 \\ 
 & 5 & 0.1008 & 0.063 & 0.1105 & 0.1692 & 0.1712 & 0.2237 \\  
 & 2 & 0.1105 & 0.115 & 0.186 & 0.2962 & 0.2673 & 0.1994 \\ 
 & 4 & 0.0828 & 0.0824 & 0.1466 & 0.2231 & 0.2192 & 0.1618 \\ 
 & 1 & 0.0884 & 0.1067 & 0.1603 & 0.2846 & 0.2231 & 0.1694 \\ \hline
UNCC & mainrun1 & 0.0242 & 0.0242 & 0.0242 & 0.05 & 0.025 & 0.0453 \\ \hline
NCSU & run1 & 0.0204 & 0.0204 & 0.0204 & 0.0577 & 0.0288 & 0.0447 \\ \hline
UNCW & run1 & 0.0007 & 0.0007 & 0.0007 & 0.0038 & 0.0019 & 0.0013 \\ \hline \hline
Baseline & BM25 & 0.1743 & 0.1703 & 0.2588 & 0.3346 & 0.3 & 0.2812 \\ \hline
\end{tabular}%
}
\caption{Performance of the participating teams on video retrieval subtask of the VCVAL task.} 
\label{tab:vr-results}
\end{table}

\begin{table}[]
\resizebox{\columnwidth}{!}{%
\begin{tabular}{l|l|l|cccc}
\hline
\textbf{n} & \textbf{Team} & \textbf{RunID} & \textbf{IoU=0.3} & \textbf{IoU=0.5} & \textbf{IoU=0.7} & \textbf{mIoU} \\ \hline
\multirow{11}{*}{1} & \multirow{2}{*}{TJUMI} & run-2 & 34.62 & 21.15 & 15.38 & 26.01 \\ 
 &  & run-1 & 40.38 & 23.08 & 17.31 & 29.2 \\ \cline{2-7} 
 & NCstate & Seahawk\_run-1 & 30.77 & 17.31 & 7.69 & 20.13 \\ \cline{2-7} 
 & \multirow{5}{*}{PolySmart} & 3 & 21.15 & 13.46 & 5.77 & 14.13 \\ 
 &  & 5 & 7.69 & 1.92 & 1.92 & 6.63 \\ 
 &  & 2 & 17.31 & 11.54 & 1.92 & 11.52 \\ 
 &  & 4 & 19.23 & 9.62 & 3.85 & 12.68 \\ 
 &  & 1 & 13.46 & 7.69 & 1.92 & 9.17 \\ \cline{2-7} 
 & UNCC & mainrun1 & 0 & 0 & 0 & 0.04 \\ \cline{2-7} 
 & NCSU & run1 & 13.46 & 7.69 & 1.92 & 10.01 \\ \cline{2-7} 
 & UNCW & run1 & 0 & 0 & 0 & 0 \\ \hline \hline
\multirow{11}{*}{3} & \multirow{2}{*}{TJUMI} & run-2 & 55.77 & 38.46 & 30.77 & 43.15 \\ 
 &  & run-1 & 55.77 & 38.46 & 30.77 & 43.6 \\ \cline{2-7} 
 & NCstate & Seahawk\_run-1 & 63.46 & 42.31 & 28.85 & 43.17 \\ \cline{2-7} 
 & \multirow{5}{*}{PolySmart} & 3 & 32.69 & 23.08 & 11.54 & 24.15 \\ 
 &  & 5 & 25 & 17.31 & 9.62 & 18.46 \\ 
 &  & 2 & 32.69 & 23.08 & 11.54 & 23.73 \\ 
 &  & 4 & 28.85 & 19.23 & 11.54 & 21.61 \\ 
 &  & 1 & 36.54 & 26.92 & 9.62 & 25.93 \\ \cline{2-7} 
 & UNCC & mainrun1 & 0 & 0 & 0 & 0.04 \\ \cline{2-7} 
 & NCSU & run1 & 13.46 & 7.69 & 1.92 & 10.01 \\ \cline{2-7} 
 & UNCW & run1 & 0 & 0 & 0 & 0 \\ \hline \hline
\multirow{11}{*}{5} & \multirow{2}{*}{TJUMI} & run-2 & 59.62 & 44.23 & 36.54 & 47.32 \\ 
 &  & run-1 & 61.54 & 46.15 & 36.54 & 47.87 \\ \cline{2-7} 
 & NCstate & Seahawk\_run-1 & 73.08 & 46.15 & 32.69 & 47.86 \\ \cline{2-7} 
 & \multirow{5}{*}{PolySmart} & 3 & 36.54 & 26.92 & 17.31 & 29.19 \\ 
 &  & 5 & 26.92 & 21.15 & 13.46 & 20.72 \\ 
 &  & 2 & 36.54 & 26.92 & 15.38 & 29.63 \\ 
 &  & 4 & 36.54 & 25 & 17.31 & 28.29 \\ 
 &  & 1 & 44.23 & 32.69 & 15.38 & 31.22 \\ \cline{2-7} 
 & UNCC & mainrun1 & 0 & 0 & 0 & 0.04 \\ \cline{2-7} 
 & NCSU & run1 & 13.46 & 7.69 & 1.92 & 10.01 \\ \cline{2-7} 
 & UNCW & run1 & 0 & 0 & 0 & 0 \\ \hline \hline
\multirow{11}{*}{10} & \multirow{2}{*}{TJUMI} & run-2 & 63.46 & 48.08 & 42.31 & 50.46 \\ 
 &  & run-1 & 63.46 & 48.08 & 42.31 & 50.46 \\ \cline{2-7} 
 & NCstate & Seahawk\_run-1 & 78.85 & 63.46 & 42.31 & 56.41 \\ \cline{2-7} 
 & \multirow{5}{*}{PolySmart} & 3 & 51.92 & 38.46 & 23.08 & 38.57 \\ 
 &  & 5 & 42.31 & 30.77 & 21.15 & 32.39 \\ 
 &  & 2 & 48.08 & 42.31 & 23.08 & 39.77 \\ 
 &  & 4 & 50 & 36.54 & 21.15 & 36.72 \\ 
 &  & 1 & 51.92 & 38.46 & 25 & 38.54 \\ \cline{2-7} 
 & UNCC & mainrun1 & 0 & 0 & 0 & 0.04 \\ \cline{2-7} 
 & NCSU & run1 & 13.46 & 7.69 & 1.92 & 10.01 \\ \cline{2-7} 
 & UNCW & run1 & 0 & 0 & 0 & 0 \\ \hline
\end{tabular}%
}
\caption{Performance of the participating teams on visual answer localization subtask of the VCVAL task.} 
\label{tab:val-results}
\end{table}

\begin{table}[]
\resizebox{\columnwidth}{!}{%
\begin{tabular}{l|l|l|ccccccc}
\hline
$\theta$ & \textbf{Team} & \textbf{RunID} & \textbf{Precision} & \textbf{Recall} & \textbf{F-score} & \textbf{IoU=0.3} & \textbf{IoU=0.5} & \textbf{IoU=0.7} & \textbf{mIoU} \\ \hline
\multirow{6}{*}{0.4} & PolySmart & LLaVA-NeXT-Video-32B-Qwen\_\_GPT4o & 31.54 &	35.06 &	32.17 &	34.87 &	30.9&	17.06 &	24.13
 \\ \cline{2-10} 
 & \multirow{5}{*}{DoshishaUzlDfki} & GPT\_meta\_prompt & 35.64 &	50.36 &	39.08	&46.22&	38.85 &	20.13 &	32.04
\\ 
 &  & chatGPT\_zeroshot\_prompt & 35.04 &	51.19	& 38.91 &	47.64 &	40.23 &	21.17 &	33.06
 \\ 
 &  & mistral\_fewshot\_prompt & 40.37	& 47.39&	41.51 &	43.5	&38.81 &	21.24 &	30.65
 \\ 
 &  & mistral\_meta\_prompt & 36.99	&42.31 &	37.56 &	39.7 &	34.62 &	20.85 &	28.11
 \\
 &  & CoSeg\_meta\_prompt & 29.65 &	19.74	&22.93 &	17.48 &	15.72	& 8.78 &	12.8
 \\  \hline \hline
\multirow{6}{*}{0.5} & PolySmart & LLaVA-NeXT-Video-32B-Qwen\_\_GPT4o & 14.98 &	17.15 &	15.55 &	17.15 &	16.17&	10.9	&13.37
 \\ \cline{2-10} 
 & \multirow{5}{*}{DoshishaUzlDfki} & GPT\_meta\_prompt & 22.32 &	30.77	&24.72 &	29.77 &	27.39 &	16.24 &	22.08
\\ 
 &  & chatGPT\_zeroshot\_prompt & 23.49	& 32.59	& 26.1	& 31.59	& 29.96 &	18.24 &	23.69
 \\ 
 &  & mistral\_fewshot\_prompt & 25.56 &	31.25 &	27.08 &	30.36 &	27.88 &	18.37 &	22.15
 \\ 
 &  & mistral\_meta\_prompt & 23.54 &	28.73 &	24.88 &	28.51 &	25.76 &	18.27 &	21.05
 \\
 &  & CoSeg\_meta\_prompt & 16.8	&11.28	 & 13.16 &	10.57 &	10.35 &	7.71 &	8.73
 \\ \hline
\end{tabular}%
}
\caption{Performance of the participating teams on QFISC task focusing on closeness and alignment of the predicted steps with the ground-truth steps. The IoU is computed by considering the overlap between ground truth and predicted captions and the extension parameter  $\lambda =3$.}
\label{tab:res-qfisc-overlap}
\end{table}

\begin{table}[h]
\resizebox{\columnwidth}{!}{%
\begin{tabular}{l|l|cccccc}
\hline
\textbf{Team} & \textbf{RunID} & \multicolumn{1}{l}{\textbf{BLEU-2}} & \multicolumn{1}{l}{\textbf{BLEU-3}} & \multicolumn{1}{l}{\textbf{METEOR}} & \multicolumn{1}{l}{\textbf{ROUGE-L}} & \multicolumn{1}{l}{\textbf{SPICE}} & \multicolumn{1}{l}{\textbf{BERTScore}} \\ \hline
PolySmart & LLaVA-NeXT-Video-32B-Qwen\_\_GPT4o & 17.04 &	11.29 &	25.42	&29.54	& 23.66 &	86.29 \\ \hline
\multirow{5}{*}{DoshishaUzlDfki} & GPT\_meta\_prompt & 19.65 &	11.72	&22.12 &	34.49	&23.84 &	86.75 \\ 
 & chatGPT\_zeroshot\_prompt & 19.88	&11.47 &	22.18 &	34.51 &	23.83&	86.69 \\ 
 & mistral\_fewshot\_prompt & 17.42 &	10.07 &	19.49 &	33.69 &	23.87 &	86.74 \\ 
 & mistral\_meta\_prompt & 15.52 &	9.07	& 18.01 &	32.17 &	22.62	&86.56 \\ 
 & CoSeg\_meta\_prompt & 4.87 &	2.05 &	11.38 &	23.15 &	17.19 &	85.55 \\ \hline
\end{tabular}%
}
\caption{Performance of the participating teams on QFISC task focusing on closeness in terms of n-gram matching and semantic similarity. The results are shown considering the threshold $\theta =0.4$.}
\label{tab:res-qfisc-ngram}
\end{table}

\begin{table}[]
\resizebox{\columnwidth}{!}{%
\begin{tabular}{l|l|ccc}
\hline
\textbf{Team} & \textbf{RunID} & \multicolumn{1}{l}{\textbf{Completeness}} & \multicolumn{1}{l}{\textbf{Accuracy}} & \multicolumn{1}{l}{\textbf{Coherenece}} \\ \hline
PolySmart & LLaVA-NeXT-Video-32B-Qwen\_\_GPT4o & 4.46 &	4.46 &	4.78\\ \hline
\multirow{5}{*}{DoshishaUzlDfki} & GPT\_meta\_prompt & 3.8 &	3.76 &	4.48 \\ 
 & chatGPT\_zeroshot\_prompt & 3.62 & 3.7 & 4.56 \\ 
 & mistral\_fewshot\_prompt & 3.16 & 3.14 & 4.31 \\ 
 & mistral\_meta\_prompt & 2.86 & 3.03 & 4.16 \\ 
 & CoSeg\_meta\_prompt & 1.74 & 1.92 & 3.33 \\ \hline \hline
 Ground Truth& NA & 4.74 & 4.76 & 4.93 \\ \hline \hline
\end{tabular}%
}
\caption{Performance of the participating teams on QFISC task focusing on human evaluation of the predicted steps.}
\label{tab:res-qfisc-human}
\end{table}


\section{Results and Discussion} 
\subsection{VCVAL}
\paragraph{VCVAL Task:} The VCVAL task consists of video retrieval and visual answer localization subtasks. We presented the results of the video retrieval subtask in Table \ref{tab:vr-results}. We reported the results in terms of MAP, R@5, R@10, P@5, P@10, and nDCG. Since the relevancy of the videos is judged in terms of multi-level judgment, we consider nDCG as the primary metric for video retrieval subtask. Team NCstate achieved the best nDCG for the video retrieval subtask with a score of $0.5738$. For the video retrieval subtask, we also developed a baseline using the BM25 lexical search approach where we used the subtitles of videos and indexed them using pyserini\footnote{\url{https://github.com/castorini/pyserini}} with default hyperparameters. We retrieved the top 10 relevant videos for each question of the test set with the built index and reported the performance in Table \ref{tab:vr-results}. The baseline achieves a strong performance compared to many of the participant's approaches for video retrieval tasks.

The participating teams' visual answer localization subtask results are reported in Table \ref{tab:val-results}. The table exhibits the detailed results with varying numbers of $n$ and multiple evaluation metrics. We consider IoU=0.7 the primary metric for this subtask as it is the most strict metric, which signifies $>=70\%$ overlap between the predicted and ground-truth visual answer segments. Team TJUMI achieved the best IoU=0.7 for the visual answer localization subtask with a score of $29.2$ ($n=1$).


\subsection{QFISC}
The results of the query-focused instructional step captioning by the participating teams are presented in Table \ref{tab:res-qfisc-overlap}, \ref{tab:res-qfisc-ngram}, \ref{tab:res-qfisc-human}. The table provides a detailed breakdown of the performance using various evaluation metrics including human evaluations on the appropriate metrics. We demonstrated the performance of the submitted runs on aligning ground truth and predicted steps in terms of precision, recall, and F-score as discussed in Section \ref{sec:qfisc-evaluation}. We also show the effect of steps similarity threshold $\theta$ on the evaluation metrics as shown in Table \ref{tab:res-qfisc-overlap}. The lenient value of $\theta$ yields better results. Table \ref{tab:res-qfisc-overlap} also provides further insight into the accuracy of the model’s predicted timestamps, showing how closely the predicted timestamps align with the actual timestamps of each generated step. This illustrates the model's effectiveness in estimating precise timing across the predicted steps.

We also computed the similarity between the predicted and ground-truth steps with multiple automatic metrics and showed the results in Table \ref{tab:res-qfisc-ngram}. Since QFISC is a generation task, it demands human evaluation due to the varying degree of model generation capacity, therefore we also computed the human scores as discussed in \ref{sec:qfisc-judgement} and provided the detailed results in Table \ref{tab:res-qfisc-human}.

\section{Conclusion}
In the overview of the TREC 2024 MedVidQA track, we discussed the tasks, datasets,
evaluation metrics, participating systems, and their performance. We evaluated the performance of the submitted runs using suitable automatic as well as manual evolution metrics. 
We hope that introducing the new tasks, developed topics/visual segments, and ground truth judgments along with the human evaluation of the generated system outputs, will be useful for fostering research toward designing video question-answering systems for healthcare needs.  

\section*{Acknowledgments}
 This research was supported by the Division of Intramural Research (DIR) of the National Library of Medicine (NLM), National Institutes of Health.  
 
\bibliographystyle{unsrt}
\bibliography{sample}

\end{document}